%% file: paper.tex
\documentclass{new_tlp}
\usepackage{mathptmx}
\DeclareMathAlphabet{\mathcal}{OMS}{cmsy}{m}{n}

\usepackage{enumitem}
\setlist{nolistsep}

\input{preamble}

  \title[Beyond strong persistence when forgetting in ASP]
        {When You Must Forget:\\ beyond strong persistence when forgetting in answer set programming}

  \author[Ricardo Gon\c calves et al.]
         {RICARDO GON\c CALVES, MATTHIAS KNORR, JO\~AO LEITE\\
         NOVA LINCS, Universidade Nova de Lisboa, Portugal
         \and STEFAN WOLTRAN\\
        TU Wien, Austria}

\newtheorem{lemma}{Lemma}[section]

\begin{document}

\label{firstpage}

\maketitle

  \begin{abstract}
    Among the myriad of desirable properties discussed in the context of forgetting in Answer Set Programming (ASP), \emph{strong persistence} naturally captures its essence. Recently, it has been shown that it is not always possible to forget a set of atoms from a program while obeying this property, and a precise criterion regarding what can be forgotten has been presented, accompanied by a class of forgetting operators that return the correct result when forgetting is possible.
However, it is an open question what to do when we have to forget a set of atoms, but cannot without violating this property.
In this paper, we address this issue and investigate three natural alternatives to forget when forgetting without violating strong persistence is not possible, which turn out to correspond to the different possible relaxations of the characterization of strong persistence. Additionally, we discuss their preferable usage, shed light on the relation between forgetting and notions of relativized equivalence established earlier in the context of ASP, and present a detailed study on their computational complexity. Under consideration for acceptance in TPLP.
  \end{abstract}


\section{Introduction}\label{sec:intro}
\input{intro}

\section{Forgetting in ASP}\label{sec:prelim}
\input{prelim}

\section{On the Limits of Forgetting}\label{sec:limits}
\input{limits}

\section{Relativized Forgetting}\label{sec:relativized}
\input{relativized}

\section{Merging $\classF_{\spF}$ and $\classF_{\AltNaive}$}\label{sec:disc}
\input{merging}

\section{Complexity}\label{sec:compl}
\input{complexity}

\section{Concluding Remarks}\label{sec:concl}
\input{conclusions}

\paragraph{Acknowledgments}
R.\ Gon\c{c}alves, M.\ Knorr and J.\ Leite were partially supported by FCT strategic project NOVA LINCS ({UID}/{CEC}/{04516}/{2013}).
R.\ Gon\c{c}alves was partially supported by FCT grant SFRH/BPD/100906/2014 and M.\ Knorr by FCT grant SFRH/BPD/86970/2012.
S. Woltran was supported by the Austrian Science Fund (FWF): Y698, P25521.

\bibliographystyle{acmtrans}
\bibliography{bib}

\end{document}

%% file: preamble.tex
\usepackage{amsmath}
\newtheorem{theorem}{Theorem}
\newtheorem{definition}{Definition}
\newtheorem{proposition}{Proposition}

\newtheorem{example}{Example}

\usepackage[normalem]{ulem}

\usepackage{color}
\definecolor{mygreen}{rgb}{0.0, 0.6, 0.0}
\definecolor{myyellow}{rgb}{1.0, 0.65, 0.0}
\usepackage{ifthen}
\newcommand{\brifnotempty}[1]{\ifthenelse{\equal{#1}{}}{}{ \br{#1}}}
\newenvironment{lemma*}[2][]
	{\pagebreak[2] \par \noindent \textbf{Lemma~\ref{#2}\brifnotempty{#1}.}\it}{\par}
\newenvironment{theorem*}[2][]
	{\pagebreak[2] \par \noindent \textbf{Theorem~\ref{#2}\brifnotempty{#1}.}\it}{\par}
\newenvironment{proposition*}[2][]
	{\pagebreak[2] \par \noindent \textbf{Proposition~\ref{#2}\brifnotempty{#1}.}\it}{\par}
\newenvironment{corollary*}[2][]
	{\pagebreak[2] \par \noindent \textbf{Corollary~\ref{#2}\brifnotempty{#1}.}\it}{\par}

\newcommand{\sign}{\mathcal{A}}

\newcommand{\htmodels}{\models_{\sf HT}}

\newcommand{\HT}{\mathcal{HT}}
\newcommand{\VHT}{\mathcal{HT}_V}
\newcommand{\classP}{\mathcal{C}}
\newcommand{\la}{\leftarrow}
\newcommand{\nf}{not\,}

\newcommand{\head}[1]{\ensuremath{\mathit{A}}}
\newcommand{\pbody}[1]{\ensuremath{\mathit{B}}}
\newcommand{\nbody}[1]{\ensuremath{\mathit{C}}}
\newcommand{\nnbody}[1]{\ensuremath{\mathit{D}}}

\newcommand{\setm}{\text{\textbackslash}}

\newcommand{\hor}{H}
\newcommand{\nor}{n}
\newcommand{\dis}{d}

\newcommand{\ex}{e}

\newcommand{\tuple}[1]{\ensuremath{\langle#1\rangle}}

\newcommand{\pE}[1]{{\bf (E$_{#1}$)}}

\newcommand{\pW}{{\bf (W)}}
\newcommand{\pPP}{{\bf (PP)}}
\newcommand{\pNP}{{\bf (NP)}}
\newcommand{\pSE}{{\bf (SE)}}
\newcommand{\pCP}{{\bf (CP)}}
\newcommand{\pSP}{{\bf (SP)}}

\newcommand{\pSI}{{\bf (SI)}}
\newcommand{\psC}{{\bf (sC)}}
\newcommand{\pwC}{{\bf (wC)}}
\newcommand{\pwE}{{\bf (wE)}}
\newcommand{\pwSP}{{\bf (wSP)}}
\newcommand{\psSP}{{\bf (sSP)}}

\newcommand{\f}[2]{\ensuremath{\mathsf{f}(#1,#2)}}
\newcommand{\fgt}{\ensuremath{\mathsf{f}}}
\newcommand{\classF}{\ensuremath{\mathsf{F}}}

\newcommand{\as}[1]{\ensuremath{\mathcal{AS}(#1)}}

\newcommand{\SRel}{\ensuremath{\mathcal{R}_{\inst}}}

\newcommand{\Rel}{\ensuremath{Rel}}
\newcommand{\RA}{\ensuremath{R}}

\newcommand{\inst}{\ensuremath{\tuple{P,V}}}

\newcommand{\smF}{{\sf SM}}

\newcommand{\naive}{\sf r}
\newcommand{\AltNaive}{\sf R}

\newcommand{\TAltNaive}{\sf M}
\newcommand{\spF}{\sf SP}

\newcommand{\Cn}{Cn}

\newcommand{\NF}{$\Omega$}

\newcommand{\op}{{\mathsf{f}}}

\usepackage{amssymb}

\newcommand{\NP}{\mbox{\rm NP}}

\newcommand{\coNP}{\mbox{\rm coNP}}
\newcommand{\SigmaP}[1]{{\Sigma}_{#1}^{P}}
\newcommand{\PiP}[1]{{\Pi}_{#1}^{P}}

%% file: intro.tex
A fundamental conclusion drawn in \cite{GoncalvesKL-ECAI16} is that it is sometimes impossible to \emph{forget} a set of atoms from an answer set program while obeying important desirable properties, notably the so-called \emph{strong persistence}. However, even in such cases, we may be forced to forget -- just imagine a court ordering the elimination of illegally acquired information. In this paper, we thoroughly investigate how to forget when it is impossible to obey strong persistence.

\emph{Forgetting} is an operation that allows the removal from a knowledge base of \emph{middle} variables no longer deemed relevant.
Its importance is witnessed by its application, e.g., to cognitive robotics \cite{LinR97,LiuW11,RajaratnamLPT14}, resolving conflicts \cite{LLM03,ZhangF06,EiterW08,LangM10}, and ontology abstraction and comparison \cite{WWTP10,KontchakovWZ10,KonevL0W12,KonevL0W13}.
With its early roots in Boolean Algebra \cite{Lewis1918}, it has been extensively studied within classical logic \cite{BledsoeH80,LLM03,Larrosa00,LarrosaMN05,MiddeldorpOI96,Moinard07,Weber86} and, more recently, in the context of Answer Set Programming (ASP)~\cite{Leite2017}. The non-monotonic rule-based nature of ASP creates very unique challenges to the development of forgetting operators -- just as it happened with other belief change operations such as revision and update, cf.\ \cite{Alferes2000,Eiter2002,Sakama2003,Slota2012,DBLP:journals/tocl/DelgrandeSTW13,DBLP:journals/tplp/SlotaL14} -- making it a special endeavour with unique characteristics distinct from those for classical logic.
This led to the introduction of several forgetting operators and classes of operators \cite{ZhangF06,EiterW08,Wong09,WangZZZ12,WangWZ13,KnorrA14,WangZZZ14,DelgrandeW15,GoncalvesKL-ECAI16} (c.f.\ \cite{GoncalvesKL16} for a recent critical survey).  

From \cite{GoncalvesKL16} it stood out that \emph{strong persistence} \pSP\ \cite{KnorrA14} -- a property essentially requiring that all existing relations between the atoms not to be forgotten be preserved -- best captures the essence of forgetting in the context of ASP. However, as shown in \cite{GoncalvesKL-ECAI16}, sometimes the atoms to be forgotten play such a pivotal role that they cannot be forgotten without violating \pSP. The class of situations when forgetting is possible was characterized through a criterion -- \NF\ -- that can be applied to any answer set program $P$ and set of atoms $V$, holding whenever $V$ cannot be forgotten from $P$, and not holding otherwise. For those cases when forgetting is possible, \cite{GoncalvesKL-ECAI16} also presents a class of forgetting operators that satisfy \pSP, dubbed $\classF_{\spF}$. 

But what if \NF\ is true and we nevertheless \emph{must} forget? This may happen for legal and privacy issues, including, for example, to enforce the new EU General Data Protection Regulation \cite{eu:gdpr}, which includes the \emph{right to be forgotten} -- the person's right to ask a corporation to eliminate private data -- or the implementation of court orders to eliminate certain pieces of illegally acquired or maintained information. Tools that can help companies and users automate the operation of forgetting should be able to handle not only situations where we can achieve the required forgetting without violating \emph{strong persistence}, but also situations where such \emph{ideal} forgetting is not possible.
Towards developing a theoretical ground on which such universally applicable tools can be based, in this paper, we thoroughly address the question of how to forget when \NF\ is true, along three different ways.

We \emph{first} take a closer look at the class $\classF_{\spF}$, which had only been considered for the case when \NF\ is false, and investigate how it behaves in general. One crucial observation is that it overestimates answer sets, i.e., forgetting preserves all existing answer sets, but new ones may be added, which indicates a violation of property \psC\ (strengthened consequence).

Our \emph{second} approach borrows from the notion of \emph{relativized equivalence} \cite{EiterFW07}, a generalization of strong equivalence that considers equivalence only w.r.t.\ a given subset of the language, and is characterized by the so-called $V$-HT-models\footnote{Programs $P_1, P_2$ are \emph{relativized equivalent} w.r.t.\ $V\subseteq \sign$ if and only if they have the same $V$-HT-models}, which lead us to consider two novel ways to forget: 
a specific operator that simply returns all rules that are relativized equivalent to the original program w.r.t.\ the atoms not to be forgotten and, alternatively, a class of operators whose result is characterized by the set of $V$-HT-models, omitting the atoms to be forgotten. The former operator turns out to be a member of the latter class. Whereas this class never overestimates answer sets, i.e., it obeys \psC, it may lose some of the original answer sets, which indicates a violation of property \pwC\ (weakened consequence).

The \emph{third} approach tries to overcome a weakness of the second, i.e., its result diverges from $\classF_{\spF}$ even when it is possible to forget, and proposes a case-based definition that can be seen as a mixture of the previous two. Whereas it preserves all answer sets, i.e., it obeys both \psC\ and \pwC, it no longer satisfies \pSI\ (strong invariance), i.e., forgetting first and then adding some set of rules $R$ (not containing the atoms to be forgotten) is no longer (strongly) equivalent to adding $R$ first, and forgetting subsequently.

In this paper, we fully investigate these three alternatives. We characterize them by showing which subset of the properties previously considered in the literature each of them obeys, study their computational complexity, and relate them by considering further additional properties to help clarify their preferable usage. Perhaps one of the most interesting features of this set of alternatives stems from a characterisation of \pSP\ according to which a forgetting operator obeys \pSP\ if and only if it obeys \psC, \pwC\ and \pSI. Hence, each of the three alternatives exactly corresponds to the relaxation of one of these three properties that characterize \pSP. 

Additional relevant results include a formal correspondence between $V$-HT-models and HT-models allowing us to leverage beneficial properties of HT-models, such as monotonicity, in the realm of $V$-HT-models, which do not satisfy them, and a complexity result for checking whether \NF\ holds.

The remainder of the paper starts with some background on forgetting in ASP, then proceeds with one section for each of the three approaches, followed by one on their complexity, and one with some brief concluding remarks.

%% file: prelim.tex
In this section, we recall the necessary notions on answer set programming and forgetting. 

\paragraph{Logic programs}
We assume a \emph{propositional signature} $\sign$, a finite set of propositional atoms\footnote{Often, the term propositional variable is used synonymously.}.
An \emph{(extended) logic program} $P$ over $\sign$ is a finite set of \emph{(extended) rules} of the form
\begin{align}
a_1 \vee \ldots \vee a_k  \la b_1,..., b_l, \nf c_{1},..., \nf c_m, \nf \nf d_1,..., \nf \nf d_n \; , \label{l:rule}
\end{align}
where all $a_1,\ldots,a_k,b_1,\ldots, b_l,c_{1},\ldots, c_m$, and $d_{1},\ldots, d_n$ are atoms of $\sign$.\footnote{Extended logic programs \cite{LifschitzTT99} are actually more expressive, but this form is sufficient here.} 
Such rules $r$ are also commonly written in a more succinct way as 
\begin{equation}
A \la B, \nf C, \nf \nf D \; , \label{l:shortRule}
\end{equation}
where we have $\head{r} = \{a_1,\ldots,a_k\}$, $\pbody{r}=\{b_1,\ldots, b_l\}$, $\nbody{r}=\{c_{1},\ldots, c_m\}$, $\nnbody{r}=\{d_{1},\ldots, d_n\}$,
and we will use both forms interchangeably. By $\sign(P)$ we denote the set of atoms appearing in $P$.
This class of logic programs, $\classP_{\ex}$, includes a number of special kinds of rules $r$: if $n=0$, then we call $r$ \emph{disjunctive}; if, in addition, $k\leq 1$, then $r$ is \emph{normal}; if on top of that $m=0$, then we call $r$ \emph{Horn}, and \emph{fact} if also $l=0$. 
The classes of \emph{disjunctive}, \emph{normal} and \emph{Horn programs}, $\classP_{\dis}$, $\classP_{\nor}$, and $\classP_{\hor}$, are defined resp.\ as a finite set of disjunctive, normal, and Horn rules. 
Given a program $P$ and a set $I$ of atoms, the \emph{reduct} $P^I$ is defined as $P^I = \{\head{r}\la \pbody{r} : r \text{ of the form (\ref{l:shortRule}) in } P, \nbody{r}\cap I=\emptyset, \nnbody{r}\subseteq I\}$.

An \emph{HT-interpretation} is a pair $\langle X,Y\rangle$ s.t.\ $X\subseteq Y \subseteq \sign$.
Given a program $P$, an HT-interpretation $\langle X,Y\rangle$ is an \emph{HT-model of $P$} if $Y\models P$ and $X\models P^{Y}$, where $\models$ denotes the standard consequence relation for classical logic.
We admit that the set of HT-models of a program $P$ are restricted to $\sign(P)$ even if $\sign(P)\subset \sign$.
We denote by $\HT(P)$ the set of \emph{all HT-models of $P$}.
A set of atoms $Y$ is an \emph{answer set} of $P$ if $\tuple{Y,Y}\in\HT(P)$, but there is no $X\subset Y$ such that $\tuple{X,Y}\in\HT(P)$.
The set of all answer sets of $P$ is denoted by $\as{P}$.
We say that two programs $P_1, P_2$ are \emph{equivalent} if $\as{P_1}=\as{P_2}$ and \emph{strongly equivalent}, denoted by $P_1\equiv P_2$, if $\as{P_1\cup R}=\as{P_2\cup R}$ for any $R\in \classP_{\ex}$.
It is well-known that $P_1\equiv P_2$ exactly when $\HT(P_1)=\HT(P_2)$~\cite{LPV01}.
We say that $P'$ is an \emph{HT-consequence} of $P$, denoted by $P\htmodels P'$, whenever $\HT(P)\subseteq \HT(P')$.
The \emph{$V$-exclusion} of a set of answer sets (a set of HT-interpretations) $\mathcal{M}$, denoted $\mathcal{M}_{\parallel V}$, is $\{X\text{\textbackslash} V\mid X\in\mathcal{M}\}$ ($\{\tuple{X\text{\textbackslash} V,Y\text{\textbackslash} V}\mid \tuple{X,Y}\in\mathcal{M}\}$). 
Finally, given two sets of atoms $X,X'\subseteq \sign$, we write $X\sim_V X'$ whenever $X\text{\textbackslash} V=X'\text{\textbackslash} V$.

We recall the notion of $A$-SE-models \cite{EiterFW07}, but here adapted to $V$-HT-models that focus on $V\subseteq \sign$, instead of on $A=\sign\text{\textbackslash} V$. 
An HT-interpretation $\tuple{X,Y}$ is called a \emph{$V$-HT-interpretation} if either $X=Y$ or $X\subset Y\text{\textbackslash} V$.
A $V$-HT-interpretation $\tuple{X,Y}$ is a \emph{(relativized) $V$-HT-model of $P$} if: 
$(a)$ $Y\models P$;
$(b)$ for all $Y'\subset Y$ with $Y\sim_V Y'$, $Y'\not\models P^Y$; and
$(c)$ if $X\subset Y$, then there exists $X'\subseteq Y$ such that $X=X'\text{\textbackslash} V$ and $X'\models P^Y$.
We denote by $\VHT(P)$ the set of \emph{all $V$-HT-models of $P$}.
Programs $P_1, P_2$ are \emph{relativized equivalent} w.r.t.\ $V\subseteq \sign$, denoted by $P_1\equiv_V P_2$, if $\as{P_1\cup R}=\as{P_2\cup R}$ for any $R\in \classP_{\ex}$ s.t.\ $\sign(R)\subseteq \sign\text{\textbackslash} V$.
We have that $P_1\equiv_V P_2$ exactly when $\VHT(P_1)=\VHT(P_2)$~\cite{EiterFW07}.

\paragraph{Forgetting}
Given a class of logic programs $\classP$ over $\sign$, a \emph{forgetting operator (over $\classP$)} is a partial function $\op:\classP\times 2^{\sign}\to \classP$ s.t. $\f{P}{V}$ is a program over $\sign(P)\text{\textbackslash} V$, for each $P\in \classP$ and $V\subseteq \sign$. 
We call $\f{P}{V}$ the \emph{result of forgetting about $V$ from $P$}.
Unless stated otherwise, in what follows, we will be focusing on $\classP=\classP_{\ex}$, and we leave $\classP$ implicit.
Furthermore, $\op$ is called \emph{closed} for $\classP'\subseteq\classP$ if, for every $P\in \classP'$ and $V\subseteq \sign$, we have $\f{P}{V}\in \classP'$.
A \emph{class $\classF$ of forgetting operators (over $\classP$)} is a set of forgetting operators (over $\classP'$) s.t.\ $\classP'\subseteq \classP$.
Such classes are usually described by a common definition/condition that each operator in the class has to satisfy (see \cite{GoncalvesKL16} for an overview on the many different kinds and forms of defining such classes).

At the same time, previous work on forgetting in ASP has introduced a variety of desirable properties accompanying these classes of operators.
In the following, we recall these properties and leave the details, e.g., on which class of forgetting operators satisfies which properties to \cite{GoncalvesKL16,GoncalvesKLJELIA16}.\footnote{We omit \pNP\ from the list, as it has been shown there to coincide with \pW.}
Unless stated otherwise, $\classF$ is a class of forgetting operators, and $\classP$ the class of programs over $\sign$ of a given $\fgt\in \classF$.
\begin{itemize}[align=left]
\item[\psC] $\classF$ satisfies \emph{strengthened Consequence} if, for each $\fgt\in \classF$, $P\in \mathcal{C}$ and $V\subseteq \sign$, we have $\as{\f{P}{V}}\subseteq \as{P}_{\parallel V}$.

\item[\pwE] $\classF$ satisfies \emph{weak Equivalence} if, for each $\fgt\in \classF$, $P, P'\in \mathcal{C}$ and $V\subseteq \sign$, we have 
$\as{\f{P}{V}}=\as{\f{P'}{V}}$ whenever $\as{P}=\as{P'}$.

\item[\pSE] $\classF$ satisfies \emph{Strong Equivalence} if, for each $\fgt\in \classF$, $P, P'\in \mathcal{C}$ and $V\subseteq \sign$: if $P\equiv P'$, then $\f{P}{V}\equiv \f{P'}{V}$.

\item[\pW] $\classF$ satisfies \emph{Weakening} if, for each $\fgt\in \classF$, $P\in \mathcal{C}$ and $V\subseteq \sign$, we have $P\htmodels\f{P}{V}$. 

\item[\pPP] $\classF$ satisfies \emph{Positive Persistence} if, for each $\fgt\in \classF$, $P\in \mathcal{C}$ and $V\subseteq \sign$: if $P\htmodels P'$, with $P'\in \mathcal{C}$ and $\sign(P')\subseteq \sign\text{\textbackslash} V$, then $\f{P}{V}\htmodels P'$. 

 \item[\pSI] $\classF$ satisfies \emph{Strong (addition) Invariance} if, for each $\fgt\in \classF$, $P\in \mathcal{C}$ and $V\subseteq \sign$, we have $\f{P}{V}\cup R \equiv \f{P\cup R}{V}$ for all programs $R\in \classP$ with $\sign(R)\subseteq \sign\text{\textbackslash} V$.

\item[\pE{\classP}] $\classF$ satisfies \emph{Existence for $\classP$}, i.e., $\classF$ is \emph{closed for a class of programs $\classP$} if there exists $\op\in \classF$ s.t.\ $\op$ is closed for $\classP$. 

\item[\pCP] $\classF$ satisfies \emph{Consequence Persistence} if, for each $\fgt\in \classF$, $P\in \mathcal{C}$ and $V\subseteq \sign$, we have $\as{\f{P}{V}}=\as{P}_{\parallel V}$.

\item[\pwC] $\classF$ satisfies \emph{weakened Consequence} if, for each $\fgt\in \classF$, $P\in \mathcal{C}$ and $V\subseteq \sign$, we have $\as{P}_{\parallel V}\subseteq \as{\f{P}{V}}$. 

\item[\pSP] $\classF$ satisfies \emph{Strong Persistence} if, for each $\fgt\in \classF$, $P\in \mathcal{C}$ and $V\subseteq \sign$, we have $\as{\f{P}{V}\cup R}=\as{P\cup R}_{\parallel V}$, for all programs $R\in \classP$ with $\sign(R)\subseteq \sign\text{\textbackslash} V$. 
\end{itemize}
We refer to the recent critical survey~\cite{GoncalvesKL16} for a discussion about existing relations between these properties, but we want to point out that the importance of \pSP\ is witnessed by the fact that if some class $\classF$ satisfies \pSP, then it also satisfies basically all other mentioned properties (but \pW\ and \pE{\classP}, which is orthogonal).

\begin{example}\label{exampleFsasNotW}
Consider the following program $P$.
\begin{align*}
a &\la \nf b &  b & \la \nf c & e & \la d & d &\la a
\end{align*}
First, if we want to forget about an atom, then we expect that all rules that do not mention this atom should persist, while rules that do mention the atoms should no longer occur.
For example, when forgetting about $d$ from $P$, the first two rules should be contained in the result of the forgetting, while the latter two should not.
At the same time, implicit dependencies should be preserved, such as, $e$ depending on $a$ via $d$.
Hence, we expect $\fgt(P,\{d\})$ as follows:
\begin{align*}
a &\la \nf b &  b & \la \nf c &  e &\la a
\end{align*}
In fact, many existing notions of forgetting in the literature (c.f.\ \cite{GoncalvesKL16}) provide precisely this result.

Now, consider forgetting about $b$ from $P$. Note that $P$ contains an implicit dependency between $a$ and $c$, namely, whenever $c$ becomes true, then so does $a$, i.e., if we add, e.g., $c\la$ to the program, then $a$ is necessarily true. 
Different notions of (classes of) forgetting operators $\fgt$ existing in the literature (see \cite{GoncalvesKL16}) would return the result of forgetting $\fgt(P,\{b\})=\emptyset$, but if we want to preserve property \pSP, then $\fgt(P,\{b\})$ must contain the rule $a\la \nf\nf c$.
In fact, a valid result for $\fgt(P,\{b\})$ such that $\fgt$ satisfies \pSP\ is:
\begin{align*}
a &\la \nf \nf c & e & \la d & d &\la a
\end{align*}
Finally, if the atom to be forgotten does not appear at the same time in some rule body and some rule head, usually no dependencies need to be preserved. Consider forgetting about $c$ from $P$, then, since $c$ only appears in the body of a rule, the result $\fgt(P,\{c\})$ is:
\begin{align*}
a &\la \nf b &  b & \la & e & \la d & d &\la a
\end{align*}
\end{example}

%% file: limits.tex

As argued in \cite{GoncalvesKL-ECAI16}, \pSP\ is the central property one wants to ensure to hold when forgetting atoms from an answer set program, essentially because its definition intuitively requires that all (direct and indirect) dependencies between the atoms not to be forgotten be preserved.
This is witnessed by the fact that any class of forgetting operators that satisfies \pSP\ also satisfies all other properties introduced in the literature, with the exception of \pW, which has been shown to be incompatible with \pSP\ \cite{GoncalvesKL16}.
However, it is also shown that it is not always possible to forget a set of atoms from a given program, that is, there is no forgetting operator that satisfies \pSP\ and that is defined for all pairs $\tuple{P,V}$, called \emph{forgetting instances}, where $P$ is a program and $V$ is a set of atoms to be forgotten from $P$.
The precise characterization of when it is not possible to forget while satisfying \pSP\ is given by means of criterion $\Omega$.

\begin{definition}[Criterion $\Omega$]
\label{def:critNF}
Let $P$ be a program over $\sign$ and $V\subseteq \sign$. 
An instance $\tuple{P,V}$ \emph{satisfies criterion \NF} if there exists $Y\subseteq \sign\text{\textbackslash} V$ such that the set of sets 
\[
\SRel^Y=\{\RA_{\tuple{P,V}}^{Y,A}\mid A\in \Rel_{\tuple{P,V}}^Y\}
\]
 is non-empty and has no least element, where 
\begin{align*}
\RA^{Y,A}_{\tuple{P,V}} & =\{X\text{\textbackslash} V\mid \tuple{X,Y\cup A}\in \HT(P)\} \\ 
 \Rel_{\tuple{P,V}}^Y & =\{A\subseteq V\mid \tuple{Y\cup A,Y\cup A}\in \HT(P) \text{ and } \\
  & \hspace{0.5cm} \nexists A'\subset A \text{ s.t.\ }\tuple{Y\cup A',Y\cup A}\in \HT(P)\}.
 \end{align*}
\end{definition}
The rationale is that each set $\SRel^Y$ is based on $Y\subseteq \sign\setm V$, which is a potential answer set of the result of forgetting.
Taking property \pSP\ into account, an answer set $Y$ of $\fgt(P,V)\cup R$ must be obtained from an answer set $Y\cup A$ of $P\cup R$, for some $A\subseteq V$.
So, the HT-models of the form $\tuple{X,Y\cup A}$ in $\HT(P)$ must be taken into account. This is captured by the set $\RA^{Y,A}_{\tuple{P,V}}$. Nevertheless, there are some $A\subseteq V$ such that $Y\cup A$ is never an answer set of $P\cup R$, for any $R$ over $\sign\setm V$. This is captured by the condition of the set $\Rel_{\tuple{P,V}}^Y$.

This criterion was shown to be sound and complete, i.e., it is not possible to forget about a set of atoms $V$ from a program $P$ exactly when $\tuple{P,V}$ satisfies criterion \NF.
A corresponding class of forgetting operators, $\classF_{\spF}$, was introduced.

\begin{definition}[SP-Forgetting]
Let $\classF_{\spF}$ be the class of forgetting operators defined by the following set:\footnote{The definition is slightly generalized from \cite{GoncalvesKL-ECAI16} as $Y$ is no longer restricted to be $Y\subseteq \sign(P)\text{\textbackslash} V$. Whenever $\sign=\sign(P)$, then the two versions naturally coincide.}
\begin{align*}
\{\fgt\mid \HT(\f{P}{V}) \! = \! \{\tuple{X,Y}\mid Y\subseteq \sign\text{\textbackslash} V \wedge X\!\in \bigcap\SRel^Y \}\}
\end{align*}
\end{definition}

It was shown that every operator in $\classF_{\spF}$ satisfies \pSP\ for instances that do not satisfy  \NF. 
In fact, restricted to those instances, $\classF_{\spF}$ satisfies every property except \pW, which makes this class of operators an ideal choice whenever forgetting is possible. 

However, the question as to whether this class is also of any use in case \NF\ is satisfied has not been tackled.
Given our focus on this problem, we first consider $\classF_{\spF}$ itself as a possible solution and characterize which of the well-known properties of forgetting are satisfied by $\classF_{\spF}$ in general, i.e., independently of whether \NF\ is satisfied or not. 
\begin{proposition}\label{prop:SPproperties}
 $\classF_{\spF}$ satisfies \pwC, \pSE, \pPP, and \pSI, but does not satisfy \pwE, \pW, \psC, \pCP.
\end{proposition}
Regarding \emph{existence}, it has already been shown in \cite{GoncalvesKL-ECAI16} that $\classF_{\spF}$ is closed for extended programs and Horn programs, but not for disjunctive nor normal programs.

From the previous proposition, we observe that \pW\ is no longer the only property that does not hold.
Notably, the fact that $\classF_{\spF}$ does not satisfy \pCP, and in particular \psC, means that there are instances $\tuple{P,V}$ for which the result of forgetting about $V$ from $P$ has answer sets that do not correspond to answer sets in the original program $P$, which is also why $\pwE$ does not hold.

\begin{example}\label{ex:addedAS}
Consider the following program $P$.
\begin{align*}
a & \la p & b & \la \nf p & p & \la \nf\nf p
\end{align*} 
Clearly, $P$ has six HT-models, $\tuple{ap,ap}, \tuple{b,b}, \tuple{b,ab}$, $\tuple{ab,ab}, \tuple{ap,abp}, \tuple{abp,abp}$\footnote{We follow a common convention and abbreviate sets in HT-interpretations such as $\{a,b\}$ with the sequence of its elements, $ab$.}, and two answer sets $\{a,p\}$ and $\{b\}$.
Intuitively, $p$ yields an exclusive choice between $a$ and $b$. 
If we take $V=\{p\}$, then, $\SRel^{\emptyset}=\emptyset$, $\SRel^{\{a\}}=\{\{a\}\}$, $\SRel^{\{b\}}=\{\{b\}\}$, and $\SRel^{\{a,b\}}=\{\{b,ab\},\{a,ab\}\}$. From this we have that $\bigcap \SRel^{\emptyset}=\emptyset$,
$\bigcap\SRel^{\{a\}}=\{a\}$, $\bigcap\SRel^{\{b\}}=\{b\}$, and $\bigcap\SRel^{\{a,b\}}=\{ab\}$. This means that for any $\fgt\in \classF_{\spF}$, $\fgt(P,V)$ has three HT-models, $\tuple{a,a}, \tuple{b,b},\tuple{ab,ab}$, which means that $\fgt(P,V)$ has three answer sets, the two from $P$ ignoring $p$, $\{a\}$ and $\{b\}$, and additionally $\{a,b\}$. 
Intuitively, this happens because using the intersection essentially discards both $\tuple{b,ab}$ and $\tuple{ap,abp}$ (modulo the forgotten $p$).
\end{example}
This is in fact rather atypical as so far no class of forgetting operators that satisfies \pwC, but not \psC, and thus not \pCP, was known. Since the violation of \psC\ may be seen as sufficient cause to render $\classF_{\spF}$ inadequate when \NF\ is satisfied -- notably when the introduction of new answer sets as the result of forgetting cannot be accepted -- alternatives need to be investigated.

%% file: relativized.tex
In this section, we explore alternative ways to forgetting in ASP, borrowing from the notion of relativized equivalence~\cite{EiterFW07}. Relativized equivalence is a generalization of strong equivalence that considers equivalence w.r.t.\ a given subset of the language, such that equivalence and strong equivalence are its special cases (for the empty and the entire language respectively).  
This fits naturally within the idea of forgetting in ASP, in particular w.r.t.\ property \pSP, inasmuch as after forgetting about $V$ from $P$ we only allow the addition of programs over $\sign\text{\textbackslash}V$, so relativized (strong) equivalence should be applied accordingly.

Based on this idea, we first define a forgetting operator that simply considers all logical consequences w.r.t.\ relativized equivalence.
This way, the result of forgetting about $V$ from $P$ amounts to the set of all rules (over $\sign\text{\textbackslash}V$) that can be added to $P$ while preserving relativized equivalence.
Given a program $P$ and $V\subseteq \sign$, we consider the closure of $P$ given $V$:
\[\Cn(P,V)=\{r \mid \{r\}\in \mathcal{C}_\ex \text{ and } P\cup\{r\} \equiv_{V}P\}.\]

Then, the result of forgetting about $V$ from $P$ is defined as
\[
\fgt_{\naive}(P,V)=\{r \mid r \in \Cn(P,V) \text{ and } \sign(\{r\})\cap V=\emptyset      \}.
\]
The resulting program does not mention the forgotten atoms and we can show that this operator does not belong to $\classF_{\spF}$.
\begin{example}\label{ex:naiveOperator}
Recall program $P$ from Ex.~\ref{ex:addedAS}.
It can be verified that $\fgt_{\naive}(P,\{p\})$ is strongly equivalent to the program:
\begin{align*}
a & \la \nf b & \bot & \la \nf a,\nf b\\
b & \la \nf a & a\vee b & \la
\end{align*}
Notably, this program does not have the answer set $\{a,b\}$, which indicates that this operator does not belong to $\classF_{\spF}$. 
\end{example}

We can show that $\fgt_{\naive}$ is well-defined, in the sense that testing relativized equivalence for each rule individually is the same as testing the entire set of rules as a whole.

\begin{proposition}\label{prop:NaiveWellDefined}
 Let $P$ be a program, $V\subseteq \sign$ and $R_1, R_2$ programs over $\sign\text{\textbackslash} V$.
 Then, $P\cup R_1\cup R_2\equiv_V P$ iff $P\cup R_1\equiv_V P$ and $P\cup R_2\equiv_V P$.
\end{proposition}
As a consequence of the above result, $\fgt_{\naive}(P,V)$ is in fact the largest set of rules over $\sign\text{\textbackslash} V$ that can be safely added to $P$ without changing its set of $V$-HT-models.

\begin{proposition}\label{prop:characterizationNaive}
Let $P$ be a program and $V\subseteq \sign$. Then, $\fgt_{\naive}(P,V)$ is the largest set of rules $R$  over the alphabet $\sign\text{\textbackslash} V$ such that $P\cup R\equiv_V P$.
\end{proposition}

We could now define a (possibly singleton) class of operators that generalizes the idea of $\fgt_{\naive}$ in a straightforward manner, and then study this class, but its definition would not be very concise, as we would always have to check for each rule whether it is relativized equivalent to the original program.

Instead, inspired by knowledge forgetting \cite{WangZZZ14}, we follow a different idea, defining a class of forgetting operators that consider the $V$-HT-models of $P$ and omit all occurrences of elements of $V$ from these.
Formally:
\[
\classF_{\AltNaive} = \{\fgt\mid \HT(\f{P}{V}) ={ \VHT(P)}_{\parallel V}\}
\]
\begin{example}
Recall Ex.~\ref{ex:naiveOperator}.
It can be verified that the result of forgetting for any $\fgt\in\classF_{\AltNaive}$ coincides with that for $\fgt_{\naive}$. 
\end{example}
It turns out that this correspondence is no mere coincidence.
In fact, we show in the following that $\fgt_{\naive}\in \classF_{\AltNaive}$, and in the course of that, we establish a precise relation between the HT-models and the $V$-HT-models of a program. 
This is an important contribution, since it allows the usage of well-known properties of HT-models, such as monotonicity, that are not satisfied by $V$-HT-models (see \cite{EiterFW07}).

First, we introduce an alternative characterization of the $V$-HT-models of a program $P$ based on its HT-models using the following notion.

\begin{definition}\label{def:relevant}
Let $P$ be a program and $Y,V\subseteq \sign$. Then, $Y$ is \emph{relevant for $P$ w.r.t.\ $V$} if
 
\begin{enumerate}[label=(\roman*)]
 
\item $\tuple{Y,Y}\in \HT(P)$

\item $\tuple{Y',Y}\notin \HT(P)$ for every $Y'\subset Y$ s.t.\ $Y\sim_V Y'$.
\end{enumerate}
$Rel(P,V)$ denotes the set of all sets relevant for $P$ w.r.t.\ $V$.
\end{definition}
This notion is tightly connected with the sets in the definition of criterion $\Omega$, i.e., we can show that $Y\cup A\in Rel(P,V)$ iff $A\in Rel^{Y}_{\tuple{P,V}}$.
This allows the alternative definition of a $V$-HT-model in terms of HT-models.

\begin{proposition}\label{prop:alternativeVmodels}
Let $P$ be a program and $V\subseteq \sign$.
Then, a $V$-HT-interpretation $\tuple{X,Y}$ is a $V$-HT-model of $P$ iff the following conditions hold:
\begin{enumerate}[label*=(\arabic*)]
 \item $Y\in Rel(P,V)$;
 
 \item If $X\subset Y$, then there exists $X'\subset Y$ with $X=X'\text{\textbackslash} V$ such that $\tuple{X',Y}\in \HT(P)$.
 \end{enumerate}
\end{proposition}

We can now present an alternative characterization of the set of $V$-HT-models of a program in terms of its set of HT-models. This result is particularly useful since it shows how the set of $V$-HT-models of a program can be directly obtained from its set of HT-models. 

\begin{proposition}\label{prop:VHTvsHT} 
Let $P$ be a program and $V\subseteq \sign$. Then,
\begin{align*}
\VHT(P) = \bigcup_{Y\in Rel(P,V)} (\{\tuple{X\text{\textbackslash} V,Y}: \tuple{X,Y}\in \HT(P)\text{ and } X\subset Y\} \cup \{\tuple{Y,Y}\}).
\end{align*}
\end{proposition}

Based on that, we can show that $\fgt_{\naive}$ is indeed a concrete forgetting operator in the class $\classF_{\AltNaive}$.
\begin{theorem}\label{theo:naiveCharac}
  Let $P$ be a program and $V\subseteq \sign$. Then,
  \[\HT(\fgt_{\naive}(P,V))_{\parallel V}=\VHT(P)_{\parallel V}.\]
\end{theorem}

Interestingly, we are also able to provide an alternative characterization of $\classF_{\AltNaive}$ that clarifies the relation to $\classF_{\spF}$.
\begin{theorem}\label{theo:classesCoincide}
Let $P$ be a program and $V\subseteq \sign$.
Then, $\classF_{\AltNaive}$ can be given by the set 
\[\{\fgt\mid \HT(\f{P}{V}) \! = \! \{\tuple{X,Y}\mid Y\subseteq \sign\text{\textbackslash} V \wedge X\!\in \bigcup\SRel^Y \}\}.\]
\end{theorem}
Thus, this notion of forgetting based on relativized equivalence differs from $\classF_{\spF}$ by considering the union of the relevant HT-models instead of the intersection, which explains the differences observed in Ex.~\ref{ex:addedAS} and \ref{ex:naiveOperator}.

Of course, whenever $\SRel^Y$ contains only one element, union and intersection coincide, which is always the case for Horn programs.

\begin{proposition}\label{prop:HornRY}
Let $P\in \classP_{\hor}$ and $V\subseteq \sign$. Then, for every $Y\subseteq \sign\text{\textbackslash} V$, we have that $\SRel^Y$ has at most one element.
\end{proposition}

Thus, when restricted to $\classP_{\hor}$, $\classF_{\AltNaive}$ coincides with $\classF_{\spF}$.

\begin{proposition}\label{prop:HornFSP=FAN1}
 Let $P\in \classP_{\hor}$ and $V\subseteq \sign$. Then, for every $\fgt\in \classF_{\spF}$ and $\fgt'\in \classF_{\AltNaive}$ we have that 
 $\fgt(P,V)\equiv\fgt'(P,V)$.
\end{proposition}

Since this correspondence does not hold in general, we also establish which properties are satisfied by $\classF_{\AltNaive}$.

\begin{proposition}\label{prop:FAN2properties}
 $\classF_{\AltNaive}$ satisfies \psC, \pSE, \pPP, \pSI, \pE{\classP_{\hor}}, \pE{\classP_{\ex}}, but not \pwE, \pW, \pwC, \pCP, \pE{\classP_{\nor}}, \pE{\classP_{\dis}}.
\end{proposition}

In terms of the considered set of properties, $\classF_{\AltNaive}$ and $\classF_{\spF}$ only differ with respect to \psC\ and \pwC. This difference, however, is crucial. Since $\classF_{\AltNaive}$ satisfies \psC, it approximates the set of answer sets of $P$, but, contrary to $\classF_{\spF}$, never ends up adding new answer sets to the result of forgetting. However, it's not all roses, as will become clear next.

%% file: merging.tex

We have shown that $\classF_{\AltNaive}$, which is based on relativized forgetting, is a better alternative than $\classF_{\spF}$ if our objective is to approximate the set of answer sets modulo the forgotten atoms, but not introduce new answer sets.
However, $\classF_{\AltNaive}$ has a drawback: there are cases where it is possible to forget while satisfying \pSP, but the result for any $\fgt\in\classF_{\AltNaive}$ does not coincide with the desired result (obtainable with operators from $\classF_{\spF}$).

\begin{example}\label{ex:hack}
Consider the following program $P$ and that we want to forget about $p$ from $P$.
\begin{align*}
a& \la p & p & \la\nf\nf p
\end{align*}
It is easy to check that $\tuple{P,V}$ does not satisfy \NF, i.e., it is possible to forget about $V$ from $P$ while satisfying \pSP. 
The result returned by any operator in $\classF_{\spF}$ is strongly equivalent to $\{a\la\nf\nf a\}$.
However, $\fgt(P,V)$ for any $\fgt\in\classF_{\AltNaive}$ is strongly equivalent to the empty program.
\end{example}

The difference between $\classF_{\spF}$ and $\classF_{\AltNaive}$, as shown in Thm.~\ref{theo:classesCoincide}, lies in the usage of intersection and union in their respective definitions.
The key point is that whenever $\SRel^Y$ has more than one element, even if there is a least one, union and intersection will not coincide. Taking this idea into account, we define a class of operators that aims at combining the delineated positive aspects of both $\classF_{\spF}$ and $\classF_{\AltNaive}$.
\begin{align*}
 \classF_{\TAltNaive}  = \{\fgt\mid \HT(\f{P}{V}) \! = \! \{\tuple{X,Y}\mid\ &Y\subseteq \sign\text{\textbackslash} V \text{ and}\\ 
 &  X\!\in \bigcup\SRel^Y \text{, if }  \SRel^Y \text{ has no least element, or}\\
 & X\!\in \bigcap\SRel^Y \text{, otherwise}\}\}.
\end{align*} 
Whenever $\SRel^Y$ has a least element, then $\classF_{\TAltNaive}$ employs the intersection, whose result is precisely the least element, similar to $\classF_{\spF}$ and does therefore coincide with the desired ideal solution in this case, and whenever there is no least element it uses the union instead, just like $\classF_{\AltNaive}$. 

\begin{example}
Consider the program of Ex.~\ref{ex:addedAS}. The result of forgetting about $p$ from that program, for any $\fgt\in\classF_{\TAltNaive}$, is strongly equivalent with that given in Ex.~\ref{ex:naiveOperator} for any $\fgt'\in\classF_{\AltNaive}$.
On the other hand, for the program given in Ex.~\ref{ex:hack}, the result of forgetting about $p$ from that program, for any $\fgt\in\classF_{\TAltNaive}$, is strongly equivalent to $\{a\la\nf\nf a\}$, and the same also holds for any operator in $\classF_{\spF}$.

\end{example}

Still, if we consider only Horn programs, then this definition of $\classF_{\TAltNaive}$ coincides with both its constituents.

\begin{proposition}\label{prop:HornFSP=FAN3}
 Let $P\in \classP_{\hor}$ and $V\subseteq \sign$. Then, for every $\fgt\in (\classF_{\spF}\cup \classF_{\AltNaive})$ and $\fgt'\in \classF_{\TAltNaive}$ we have that  $\fgt(P,V)\equiv\fgt'(P,V)$.
\end{proposition}

Moreover, unlike $\classF_{\AltNaive}$, we are able to show that, whenever it is possible to forget, $\classF_{\TAltNaive}$ coincides with $\classF_{\spF}$.

\begin{proposition}\label{prop:notOmegaFSP=FAN3}
 Let $P$ be a program and $V\subseteq \sign$, such that $\tuple{P,V}$ does not satisfy $\Omega$.
 Then, for every $\fgt\in \classF_{\spF}$ and $\fgt'\in \classF_{\TAltNaive}$ we have that 
 $\fgt(P,V)\equiv\fgt'(P,V)$.
 \end{proposition}

The particular definition of $\classF_{\TAltNaive}$ ensures that yet again a different set of properties is satisfied by it.
\begin{proposition}\label{prop:FAN3properties}
 $\classF_{\TAltNaive}$ satisfies \psC, \pwE, \pSE, \pwC, \pCP, \pPP, \pE{\classP_{\hor}}, \pE{\classP_{\ex}}, but not \pW, \pSI, \pE{\classP_{\nor}}, \pE{\classP_{\dis}}.
\end{proposition}
Contrary to $\classF_{\spF}$ and $\classF_{\AltNaive}$, the class $\classF_{\TAltNaive}$ satisfies both \pwC\ and \psC, and consequently \pCP. 
Therefore, the result of forgetting according to $\classF_{\TAltNaive}$ preserves the answer sets of $P$, but, unlike the other two, no longer satisfies \pSI.

In fact, the answer sets are no longer preserved if a (non-empty) program over $\sign\text{\textbackslash}V$ is added to $P$.
To capture this in a more precise way, we introduce generalizations of \pwC\ and \psC, which correspond to the two inclusions of \pSP.

\begin{itemize}[align=left]
\item[\psSP] $\classF$ satisfies \emph{strengthened Strong Persistence} if, for each $\fgt\in \classF$, $P\in \mathcal{C}$ and $V\subseteq \sign$, we have $\as{\f{P}{V}\cup R}\subseteq\as{P\cup R}_{\parallel V}$, for all $R\in \classP$ with $\sign(R)\subseteq \sign\text{\textbackslash} V$.
\item[\pwSP] $\classF$ satisfies \emph{weakened Strong Persistence} if, for each $\fgt\in \classF$, $P\in \mathcal{C}$ and $V\subseteq \sign$, we have $\as{P\cup R}_{\parallel V}\subseteq \as{\f{P}{V}\cup R}$, for all $R\in \classP$ with $\sign(R)\subseteq \sign\text{\textbackslash} V$. 
\end{itemize}
Property \pwSP\ guarantees that all answer sets of $P$ are preserved when forgetting, no matter which rules $R$ over $\sign\text{\textbackslash} V$ are added to $P$, but, for some such $R$, does not prevent that the result of forgetting has more answer sets than $P$. 
Vice versa, \psSP\ does not guarantee the preservation of all answer sets of $P$ for some added $R$ over $\sign\text{\textbackslash} V$, but it ensures that all answer sets of the result of forgetting indeed correspond to answer sets of $P$, independently of the added rules $R$.

We can show that each of the three considered classes of forgetting operators only satisfies one of the two properties.

\begin{theorem}\label{theo:weakstrongSP}
 $\classF_{\spF}$ satisfies \pwSP, whereas $\classF_{\AltNaive}$ and $\classF_{\TAltNaive}$ satisfy \psSP.
\end{theorem}

Since there is no class of forgetting operators that satisfies \pSP~\cite{GoncalvesKL-ECAI16}, it is clear that $\classF_{\spF}$ does not satisfy \psSP, and that $\classF_{\AltNaive}$ and $\classF_{\TAltNaive}$ do not satisfy \pwSP.
Thus, even though $\classF_{\AltNaive}$ satisfies \pwC, i.e., \pwSP\ for an empty $R$, it does not for arbitrary $R$'s.
Still, although both $\classF_{\AltNaive}$ and $\classF_{\TAltNaive}$ satisfy \psSP, the following result shows that $\classF_{\TAltNaive}$ provides a better approximation in terms of property \pSP.

\begin{proposition}\label{prop:BetterApprox}
 Let $P$ be a program, $V\subseteq \sign$, $\fgt\in \classF_{\AltNaive}$, and $\fgt'\in \classF_{\TAltNaive}$. Then,
for every $R\in \classP$ with $\sign(R)\subseteq \sign\text{\textbackslash} V$,
\[\as{\fgt(P,V)\cup R}\subseteq \as{\fgt'(P,V)\cup R}.\]
\end{proposition}

Clearly, $\classF$ satisfies \pSP\ iff it satisfies \pwSP\ and \psSP.
Since no $\classF$ can in general satisfy \pSP, we basically obtain two kinds of relaxations on the conditions of \pSP.
But we can do even better: following results from \cite{GoncalvesKL16}, we know that $\classF$ satisfies \pSP\ iff it satisfies \pwC, \psC, and \pSI. 
From the results in Props.~\ref{prop:SPproperties}, \ref{prop:FAN2properties}, and \ref{prop:FAN3properties}, we obtain that each of the three discussed classes corresponds to a unique relaxation of the conditions of \pSP, implying that our study gives a complete account on which forgetting operators to use when \pSP\ cannot be satisfied, but only approximated. 

Arguably, $\classF_{\TAltNaive}$ is also more flexible in situations where we have to forget several atoms for which $\classF_{\spF}$ and $\classF_{\AltNaive}$ do not provide the optimal overall choice.

\begin{example}
Consider the following program $P$ from which we want to forget about $c$ and $p$.
\begin{align*}
d & \la c & c & \la \nf\nf c & a & \la p & b & \la \nf p & p & \la \nf\nf p 
\end{align*}
Clearly, $\classF_{\spF}$ allows us to correctly capture the result of forgetting about $c$, in the sense that $d\la \nf\nf d$ is part of the result of forgetting, but, at the same time, will introduce new answer sets in which both $a$ and $b$ are true.
On the other hand, $\classF_{\AltNaive}$ will avoid the latter problem, but will simply cancel all rules mentioning $d$ and $c$.
Here, $\classF_{\TAltNaive}$ certainly provides the best alternative as it avoids both problems and provides the desired result.
\end{example} 

In practice, the choice between the three classes greatly depends on the application at hand. 
To help making this decision, we now identify, for each of the three classes, a set of conditions in favor of its choice over the other two.

The class $\classF_{\spF}$ should be chosen whenever:
\begin{itemize}

\item[--] \pSP\ should hold for those instances that do not satisfy $\Omega$;

\item[--] Rules that do not mention atoms to be forgotten should be preserved;

\item[--] All answer sets should be preserved; and

\item[--] We do not mind the appearance of new answer sets.

\end{itemize}

The class $\classF_{\AltNaive}$ should be chosen whenever: 
\begin{itemize}

\item[--] Rules that do not mention atoms to be forgotten should be preserved;

\item[--] No new answer sets should appear;

\item[--] We do not mind that some answer sets may disappear; and

\item[--] We do not mind that \pSP\ does not hold even if $\Omega$ does not hold.

\end{itemize}

The class $\classF_{\TAltNaive}$ should be chosen whenever: 
\begin{itemize}

\item[--] \pSP\ should hold for those instances that do not satisfy $\Omega$;

\item[--] Answer sets should be preserved precisely (modulo the forgotten atoms); and

\item[--] We do not mind to change rules that do not mention atoms to be forgotten.

\end{itemize}

These conditions stem from the obtained results on which properties each of the classes of forgetting operators satisfies, and can be seen as a guideline for a more informed choice between the three alternative classes of operators.

%% file: complexity.tex

We assume familiarity with standard complexity concepts, such as $\NP$.
Given a complexity class $\mathcal{C}$,
a $\mathcal{C}$ oracle decides a given sub-problem from $\mathcal{C}$ in one computation step.
The class $\SigmaP{k}$
contains the problems that can be
decided in polynomial time by a non-deterministic
Turing machine
with unrestricted access to a $\SigmaP{k-1}$ oracle.
$\PiP{k}$ is the complementary class of $\SigmaP{k}$.
Thus, $\SigmaP{1}=\NP$, and $\PiP{1}=\coNP$.
We also recall that a language is in complexity class $D^P_i$ iff it is the intersection of a language in $\SigmaP{i}$ and a language in $\PiP{i}$. Instead of $D^P_1$ we use the more common name $D^P$.
In addition, the following result will be useful due to the established correspondence between HT-and $V$-HT-models in Prop.~\ref{prop:VHTvsHT}.
\begin{proposition}[\cite{EiterFW07},Theorem 6.12.]\label{prop:ht}
Given a program $P$, an HT-interpretation $\tuple{X,Y}$, and $V\subseteq \sign$, deciding
whether $\tuple{X,Y}\in\VHT(P)$ is $D^P$-complete.
\end{proposition}

Our first result is in the spirit of model-checking.

\begin{lemma}\label{lemma:modelcheck}
Given program $P$, $V\subseteq \sign$, and 
HT-interpretation $\tuple{X,Y}$. Deciding
whether 
$\tuple{X,Y}\in
\VHT(P)_{\parallel V}$ is $\SigmaP{2}$-complete. 
Hardness holds already 
for disjunctive programs.
\end{lemma}
Membership follows from guessing an interpretation $Y'\sim_V Y$ and checking $(X,Y')\in\VHT(P)$ (cf.\ Proposition~\ref{prop:ht}), while the hardness result can be adapted from the $\SigmaP{2}$-hardness of ASP consistency, cf.\ \cite{EiterG95}.
By means of this, we can determine the complexity of deciding whether a given program is strongly equivalent to the result of forgetting obtained by any $\fgt\in\classF_{\AltNaive}$.

\begin{theorem}\label{thm:comp:sp1}
Given programs $P$, $Q$, and $V\subseteq \sign$, deciding 
whether 
$P\equiv \f{Q}{V}$ 
(for $\fgt\in\classF_{\AltNaive}$)
is 
$\PiP{3}$-complete.
Hardness holds already 
for disjunctive programs.
\end{theorem}
Essentially, for the complementary problem, we guess an HT-interpretation $\tuple{X,Y}$ and check that either $(X,Y)\in\HT(P)$ or $(X,Y)\in\VHT(P)_{\parallel V}$, but not both.
The hardness result is then obtained by a reduction from $(3,\forall)$-QSAT.

The next result provides the complexity of determining whether some $X$ occurs in the intersection of $\SRel^Y$ used in the definition of $\classF_{\spF}$, $\classF_{\TAltNaive}$ and \NF.

\begin{lemma}\label{lemma:modelcheck:sp}
Given program $P$, $V\subseteq \sign$, and 
HT-interpretation $\tuple{X,Y}$ with $Y\subseteq \sign\setminus V$, deciding
whether 
$
X\in \bigcap\SRel^Y
$
is in $D^P_2$.
\end{lemma}
Basically, we have to perform a $\SigmaP{2}$- and a $\PiP{2}$-test.
The former decides whether $\SRel^Y \neq\emptyset$, while the latter determines that for all 
$A\subseteq V$, 
either
$\tuple{Y\cup A,Y\cup A}\notin\VHT(P)$ or 
$\tuple{X,Y\cup A}\in\VHT(P)$. 

This Lemma allows us to obtain an identical result to Thm.~\ref{thm:comp:sp1} for $\classF_{\spF}$.

\begin{theorem}\label{thm:spf}
Given programs $P$, $Q$, and $V\subseteq \sign$, deciding 
whether 
$P\equiv \f{Q}{V}$ 
(for $\fgt\in\classF_{\spF}$)
is 
$\PiP{3}$-complete.
Hardness holds already 
for disjunctive programs.
\end{theorem}
The basic proof idea is very similar to the one sketched for Thm.~\ref{thm:comp:sp1}, but subsituting the test $(X,Y)\in\VHT(P)_{\parallel V}$ with $(X,Y)\in\HT(\fgt(P,V))$ for $\fgt\in\classF_{\spF}$.

Since the definition of $\classF_{\TAltNaive}$ is based on cases, deciding whether its condition holds, is computationally more expensive than the previous two (in Lemmas~\ref{lemma:modelcheck} and \ref{lemma:modelcheck:sp}).
\begin{lemma}\label{lemma:modelcheck:3}
Given program $P$, $V\subseteq \sign$, and 
HT-interpretation $\tuple{X,Y}$ with $Y\subseteq \sign\setminus V$, deciding
whether 
$
X\!\in \bigcup\SRel^Y \text{ if }  \SRel^Y \text{ has no least element, and } X\!\in \bigcap\SRel^Y \text{ otherwise}$,
is in $\SigmaP{3}$ and in $\PiP{3}$.
\end{lemma}

Fortunately though, since this test is both in $\SigmaP{3}$ and in $\PiP{3}$, in the next result, we can basically solve the complementary problem of guessing an HT-interpretation $\tuple{X,Y}$ and check that either $(X,Y)\in\HT(P)$ or $(X,Y)\in\HT(\fgt(P,V))$ for $\fgt\in\classF_{\TAltNaive}$, but not both, in one step.

\begin{theorem}\label{thm:comp:sp3}
Given programs $P$, $Q$, and $V\subseteq \sign$, deciding 
whether 
$P\equiv \f{Q}{V}$ 
(for $\fgt\in\classF_{\TAltNaive}$)
is 
$\PiP{3}$-complete.
Hardness holds already 
for disjunctive programs.
\end{theorem}
Thus, determining whether $P\equiv \f{Q}{V}$ for $\fgt$ of any of the three considered classes of forgetting operators is always $\PiP{3}$-complete.
This shows that the choice which of the three classes of forgetting operators to use in a concrete situation is not influenced by their computational complexity.   

Finally, we provide the complexity result for criterion \NF, which on 
the one hand improves on a flaw for the membership result in 
\cite{GoncalvesKL-ECAI16}, but also includes the hardness result here.

\begin{theorem}\label{thm:omegacomplexity}
Let $P$ be a program over $\sign$ and $V\subseteq \sign$.
Deciding whether $\tuple{P,V}$ satisfies criterion \NF\ is
$\SigmaP{3}$-complete.
Hardness holds already 
for disjunctive programs.
\end{theorem}

%% file: conclusions.tex
We addressed the problem of forgetting in ASP when we \emph{must} forget, even if satisfying the fundamental desirable property \pSP\ is not possible. 

We thoroughly investigated three alternatives which, despite stemming from different starting points -- one reusing a known class of forgetting operators, one exploring the concept of relativized equivalence, and one trying to get the best of the previous two -- turn out to each correspond to the relaxation of one of three properties -- \pwC, \psC\ and \pSI\ -- that together characterize \pSP. We characterized the three classes by showing which of the usually considered properties each obeys, established links between them, and investigated their computational complexity. 
The computational complexity turns out to be high, which is not surprising given, for example, the fact that, in classical logic, forgetting can only be expressed as a second-order axiom. Nevertheless, on the one hand, forgetting is an operation not expected to be done as regularly as for example model computation or query answering, while, on the other hand, at least for those classes that satisfy \pSI, $\classF_{\spF}$ and $\classF_{\AltNaive}$, we can perform forgetting in a modular way focusing only on the relevant part of the program.
Whether this can be extended also to $\classF_{\TAltNaive}$ remains an interesting open problem for future research.

We also established relevant novel results concerning a correspondence between $V$-HT-models and HT-models and a full complexity result for checking whether the criterion (\NF) that indicates whether it is possible to forget while satisfying \pSP\ holds.

It is also noteworthy that none of the other operators and classes of operators mentioned in the literature satisfy the properties satisfied by the three classes discussed in this paper (c.f.\ \cite{GoncalvesKL16}). The closest approximation is the operator $\classF_{\smF}$ \cite{WangWZ13} which obeys the same set of properties previously found in the literature as $\classF_{\TAltNaive}$, yet, unlike $\classF_{\TAltNaive}$, it does not satisfy either of the inclusions of \pSP, notably \psSP.

Avenues for future research include investigating different forms of forgetting which may be required in practice, such as those that preserve some aggregated meta-level information about the forgotten atoms, or even going beyond maintaining all relationships between non-forgotten atoms which may be required by certain legislation. 
This may also be of interest for semantics different from ASP, such as for forgetting under the well-founded semantics \cite{AlferesKW13,KnorrA14}.